# Deep learning method to remove chemical, kinetic and electric artifacts on ISEs

Byunghyun Ban
Future Farming Team
**Imagination Garden Inc.**
Andong Si, Republic of Korea
halfbottle@sangsang.farm

*Abstract*— I suggest a deep learning based sensor signal processing method to remove chemical, kinetic and electrical artifacts from ion selective electrodes' measured values. An ISE is used to investigate the concentration of a specific ion from aqueous solution, by measuring the Nernst potential along the glass membrane. However, application of ISE on a mixture of multiple ion has some problem. First problem is a chemical artifact which is called ion interference effect. Electrically charged particles interact with each other and flows through the glass membrane of different ISEs. Second problem is the kinetic artifact caused by the movement of the liquid. Water molecules collide with the glass membrane causing abnormal peak values of voltage. The last artifact is the interference of ISEs. When multiple ISEs are dipped into same solution, one electrode's signal emission interference voltage measurement of other electrodes. Therefore, an ISE is recommended to be applied on single-ion solution, without any other sensors applied at the same time. Deep learning approach can remove both 3 artifacts at the same time. The proposed method used 5 layers of artificial neural networks to regress correct signal to remove complex artifacts with one-shot calculation. Its MAPE was less than 1.8% and $R^2$ of regression was 0.997. A randomly chosen value of AI processed data has MAPE less than 5% (p-value 0.016).

*Keywords—AI, Machine Learning, ISE, Analog Signal Processing, Horticulture, Aqua Culture*

## I. INTRODUCTION

Convergence of IT with horticulture is not surprising anymore. Propagation of farming technology is the main trend of agricultural industry around the world. The first approach trend was to precise control of farming environment. It started from basic IoT systems in 10 years ago, but AI based facility control is not special today. [1]

The trend of automation of human labors is now changing into automation of knowledge and information. Traditional cultivation methods highly depend on the experience of a farmer. Nowadays, the replacement of experience based know-how is main trend of Farm-tech area. [2, 3]

On fertilization area, the hottest application is hydroponics aqua culture). Open hydroponic system does not recycle the nutrient solution while closed hydroponic system does. Closed

This work was supported by Agriculture, Food and Rural Affairs Convergence Technologies Program for Educating Creative Global Leader funded by the Ministry of Agriculture, Food and Rural Affairs (MAFRA, Korea). [Project Number 1545020852].

system is very popular around the world today [4], because opened system has the possibility of pollution such as eutrophication.

Recently, many researchers have suggested an ISE based hydroponic approaches in order to measure individual ion concentrations rather than EC or TDS. [5-7]

An ISE measures concentration of a specific ion by acquisition of membrane voltage, induced by target ions. The Nernst potential E along a thin membrane is defined as equation $RT/_{ZF} \ln([ion_{out}]/[ion_{in}])$, where R is gas constant, T is temperature, z is the number of ions which pass through the membrane, and F is faraday constant. This nature of an ion selective electrode causes 3 kinds of artifacts.

*A. Ion Interference Effect*

Nernst potential is induced by ion-exchange through the glass membrane with small hole which is designed for a specific target ion. This phenomenon is interfered by other ions. This artifact is described with Nikolsky-Eisenman Equation [8], provided as equation (1), where a is the thermodynamic activity of ion, $k_i$ is the selectivity coefficient and $z_i$ is the exchange of interfering ion i through glass membranes.

$$E = E_0 + \frac{RT}{zF} \ln\left[a + \Sigma_i \left(k_i a_i^{\frac{z}{z_i}}\right)\right] \quad (1)$$

A researcher needs to know the thermodynamic activity parameter $a_i$ and selectivity coefficient $k_i$ for every ion involved in the solution to predict the magnitude of this artifact. Parameter a is defined as equation (2), where $\mu$ is molar chemical potential, $\mu^\theta$ is the chemical potential under some standard condition.

$$a_i = \exp(\frac{\mu_i - \mu_i^\theta}{RT}) \quad (2)$$

The activity value changes along temperature. Therefore, measurements of molar chemical potential in not-standard condition for all individual ions are required to predict the magnitude of ion interference effect, with equation (1).

| Single  | multiple of the concentration of the original Yamazaki's solution | | | | | | | | | |
|---|---|---|---|---|---|---|---|---|---|---|
| Solvent | 0.990 | 1.961 | 2.913 | 3.846 | 4.762 | 5.660 | 6.542 | 7.407 | 8.257 | 9.091 |
| $K^+$ | 3.8835 | 7.3273 | 10.6689 | 13.8696 | 16.8848 | 19.7466 | 22.4665 | 25.0548 | 27.5208 | 29.8730 |
| $Ca^{++}$ | 0.9709 | 1.8318 | 2.6721 | 3.4674 | 4.2212 | 4.9366 | 5.6166 | 6.2637 | 6.8802 | 7.4683 |
| $NO_3^-$ | 5.8252 | 10.9910 | 16.0328 | 20.8044 | 25.2371 | 29.6199 | 33.6998 | 37.5822 | 41.2813 | 44.8096 |
| $NH_4^+$ | 0.4854 | 0.9159 | 1.3361 | 1.7337 | 2.1106 | 2.4683 | 2.8083 | 3.1312 | 3.4401 | 3.7341 |

Mol / mL

| Mixture | multiple of the concentration of the original Yamazaki's solution | | | | | | | | | |
|---|---|---|---|---|---|---|---|---|---|---|
| | 0.971 | 1.887 | 2.752 | 3.571 | 4.348 | 5.085 | 5.785 | 6.452 | 7.087 | 7.692 |
| $K^+$ | 3.9604 | 7.8431 | 11.6505 | 15.3845 | 19.0476 | 22.6415 | 26.1682 | 29.6296 | 33.0275 | 36.3636 |
| $Ca^{++}$ | 0.9901 | 1.9608 | 2.9126 | 3.8462 | 4.7619 | 5.6604 | 6.5421 | 7.4074 | 8.2569 | 9.0909 |
| $NO_3^-$ | 3.9604 | 7.8431 | 11.6505 | 15.3846 | 19.0476 | 22.6415 | 26.1682 | 29.6296 | 33.0275 | 36.3636 |
| $NH_4^+$ | 0.4951 | 0.9804 | 1.4563 | 1.9230 | 2.3810 | 2.8302 | 3.2710 | 3.7037 | 4.1284 | 5.4546 |

Mol / mL

**Table 1. Experimental Conditions**

This is not feasible in agricultural applications because most solutions used in crop cultivation area contains a variety of ions. For example, Yamazaki's solution, which is the simplest nutrient solution for leafy vegetable, has 23 different ions. [9] And the total ion composition of water used to make fertilizer should be known.

*B. Kinetic Artifacts*

Movement of solution physically interferes the glass membrane of electrode. Therefore, ISEs are recommended to be applied on a non-flowing solution and to wait around 1~10 minutes before measurement to make electrode stable. Even a drop of water make inaccurate measurement.

*C. Electrode Interference*

If two or more ISEs are applied on the same solution, an electrode's electrical signal may interfere other sensors. This artifact is a problem of real-time measurement.

Liberti et al. provided readjustment method with Gran's plots to remove ion interference effect but it also require prior knowledge of ions. [10] I suggested a machine learning approach to remove the ion interference effect performing 91.5~97.8% accuracy on test cases [11], but failed to remove the artifact from movement of solution. A distorted signal caused by drop of water made the regression fail. Recently, Woo-jae Cho et al. suggested a deep learning approach to calibrate ISE signal which showed better performance than traditional 3 pointed calibration. [4] However, the experiment was conducted with living plant whose nutrient solution is thin. My previous work was performed from that concentration, to 10 times thicker solution. As the ion interference effect becomes greater in higher concentration, Cho's work need additional verification on thicker solution.

II. ISE MEASUREMENT

An experiment with an array of ISEs was performed to acquire training/test data and to figure out the magnitude of ion interference effect.

*A. Equipments*

Vernier's Go Direct® ISE series were used to measure the concentration of ions. $K^+$, $Ca^{++}$, $NO_3^-$, $NH_4^+$ ions are measured with corresponding ion selective electrodes. The sensors were dipped into the same solution at the same time and connected to same computer, sharing same power source to remove artifacts from electricity supply.

*B. Chemical Solutions*

I used Yamazaki's nutrient solution for lettuce [9] because it is very popular around the world and its chemical phenomenon and interactions of every solvent are fully understood. [2]

But I did not applied some ions whose population is very little. $KNO_3$, $Ca(NO_3)_2$-$4H_2O$ and $NH_4H_2PO_4$ were used to make the nutrient solution. $H_xPO_4^{3-x}$ families builds complex sediments reactions but not measurable with ISEs.

I prepared the solution with 10 different concentrations. I set the concentration of the original Yamazaki's solution as 1. The experiment condition is described on Table 1.

*C. Method*

Two measurement experiments were conducted. For each experiment, ISEs are dipped into a 1 L beaker filled with distilled water at first. Then droplets of highly concentrated

| Ion | a | b | R² | Data size |
|---|---|---|---|---|
| $K^+$ | 4e-08 | 5.4996 | 0.957 | 20,590 |
| $Ca^{++}$ | 2e-15 | 16.697 | 0.9631 | 16,379 |
| $NO_3^-$ | 73.727 | -5.748 | 0.965 | 26,042 |
| $NH_4^+$ | 2e-07 | 4.8642 | 0.9441 | 26,045 |

**Table 2. Regression result for calibrations**

| MSE | MAPE | R² |
|---|---|---|
| 2.000e-5 | 56.460% | 0.879 |

**Table 3. Error analysis**

chemical solution is added to the beaker in order to induce the kinetic artifact.

For single solvent experiment for ISE calibration, only one solvent such as $KNO_3$ was taken into the solution to avoid ion interference effect and electrode interference.

During the ion mixture experiment, multiple ISEs measured the concentration at the same, taken into one beaker together, to induce ion interference effect.

Both processes were repeated 10 times.

### D. Calibration

For each ions, exponential regression on whole data was performed to calibrate the sensors because the Nernst potential is induced by the log of concentration. Each ISE was calibrated with 10-theoretically different concentration, with data point more than 10 thousands.

Calibration equation format is described as equation (3), where $C_i$ is the ISE-measured concentration, $V$ is the membrane voltage from ISE, and a and b are parameters for regression. The results of regressions are provided on Table 2.

$$C_i = a\, e^{b\,V} \quad (3)$$

### E. Artifacts

Applying the calibration equation (3) with coefficients from Table 2, the result from ion mixture experiment were

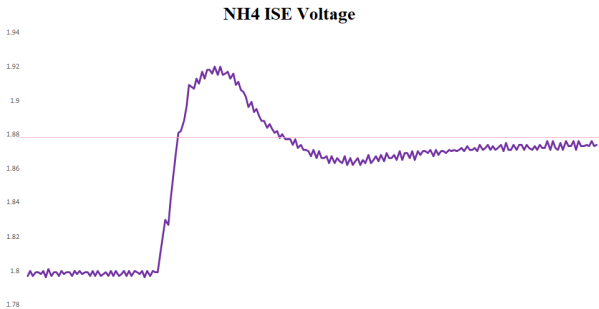

**Figure 1. Kinetic Artifact** The kinetic artifact caused by droplet is measured. This picture is taken from a part of $NH_4^+$ ISE's measurement result. Purple line is measured voltage and pink line is theoretical value. The measured kinetic artifacts tended to distort the voltage graph similar to the transient response.

| Model Number | Model Arcitecture | MAPE |
|---|---|---|
| 1 | 128-128-128-4 | 7.885 |
| 2 | 128-128-128-128-4 | 7.745 |
| 3 | 256-256-128-64-4 | 5.371 |
| 4 | 256-256-256-128-4 | 5.662 |
| **5*** | **256-256-256-256-4** | **1.779** |

* Suggested Model

**Table 4. Regression result for calibrations**

analyzed. Error between theoretical value and ISE-measured values are evaluated on Table 3. MAPE score is especially larger than other measurement.

Kinetic artifacts measured from the experiment is described in Figure 1.

### III. METHOD

#### A. Equipments and Library

NVIDA's GPU GTX1080TI (11 Gb VRAM) was used with KERAS (tensorflow backend). The source code of the proposed method is available at the author's Github repository (https://github.com/needleworm/ion_interference)

#### B. Data

Experiment data from mixture of ions is used. 20% of the data was randomly selected to be used as test data, and the others are used as training data.

Just a scale normalization was done on the experimental data before feeding into the neural network. The whole data was divided with the maximum value of the data. That value was stored on the memory and multiplied on the output values of the neural networks. Any other normalization method is not applied to train a robust model.

#### C. Objective Function & Optimization

Mean absolute percentage error(MAPE) was used as loss function. MAPE is defined as equation (4). $Y_{gt}$ is ground truth data and $Y_{predict}$ is model-predicted data.

$$\text{MAPE} = \frac{100}{n}\sum \left| \frac{Y_{gt} - Y_{predict}}{Y_{gt}} \right| \% \quad (4)$$

I avoided squared error based approach such as mean squared error because the scale of error would become too small, yielding small gradient.

ADAM optimizer [12] was used to optimize the weight parameters. Initial learning rate was 1e-4 but I also applied learning rate decay as 1e-7 per epoch. Because the model seemed to converge fast at first but failed to find sophisticated convergence point.

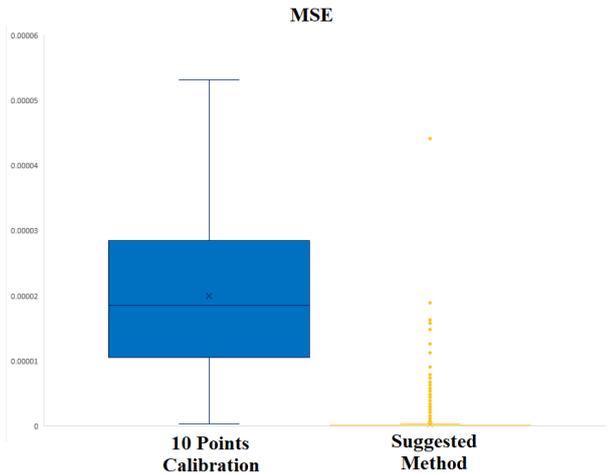

**Figure 2. MSE Box Plot** Mean square errors between predicted values and theoretical values are measured.

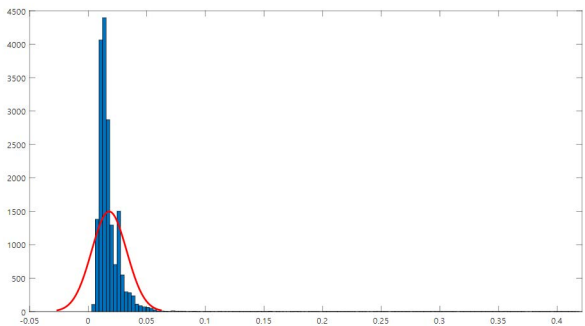

**Figure 3. MAPE Histogram with Fitting Line**

*D. Neural Network Models*

The structures of models used for experiment is described on Table 4. MAPE values at convergence state are also provided. All layers are simple fully connected networks. ReLu function and batch normalization [13] are applied on every layer except the output layer. The activation function of output layer was sigmoid function because the training and test data are normalized into a value between 0 ~ 1.

## IV. EXPERIMENT

The models were trained with minibatch size 64. After training, test data was directly fed into the model to make artifact-removed data. The weight variable and prediction result files are also provided with the source code, at the author's Github repository.

## V. RESULT

Results are summarized on Table 5.

My previous work with quadratic regression model [11] was applied on the experiment data for comparison. It failed to remove artifacts and even increased error dramatically. The value of MSE was 1.004e-3, $R^2$ was -5.092 and MAPE was 321.79%. Concerning the MAPE values on Table 4, models with higher variance with lower bias performed better regression result. It implies that quadratic model's variance was too small to regress the experimental data with 3 different artifacts. It worked well on static and stable condition, but the result shows that it is not appropriate for real-time measurement of chemical solutions which has kinetic artifacts.

| Method | MSE | $R^2$ | MAPE (%) |
|---|---|---|---|
| 10 Point Calibration | 2.00e-5 | 0.879 | 56.458 |
| Quadratic Model [11] | 1.00e-3 | -5.092 | 321.786 |
| **Proposed Method*** | **1.85e-7** | **0.997** | **1.779** |

**Table 5. Experiment Summary**

Comparison between 10 point calibration result and suggested method are plotted on Figure 2. The MAPE score of suggested method is plotted on Figure 3. It showed mean 1.17881% with standard deviation 0.014943. The hypothesis that any randomly chosen value has MAPE less than 0.05 had p-value 0.016.

## VI. CONCLUSION

A deep learning approach with MAPE loss to remove multiple artifact on ISEs was introduced. Measurement with ion selective electrodes on a chemical solution which has different ions has 3 major artifacts: ion interference effect, kinetic artifact and electrode interference. Proposed method successfully removed them with a single scalar regression task. Not only removing 3 artifacts, this model also performed a calibration to convert voltage into molarity. Previous method with quadratic regression model is not suitable to remove those 3 artifacts at the same time but proposed method removed 3 different artifacts and restored the data with mean absolute percentage error less than 1.8%. For a real-time measurement of complex solution, I recommend a high-variance model such as deep neural networks remove multiple artifacts rather than simple regression models.

ACKNOWLEDGMENT *(Heading 5)*

This work was supported by Agriculture, Food and Rural Affairs Convergence Technologies Program for Educating Creative Global Leader funded by the Ministry of Agriculture, Food and Rural Affairs (MAFRA, Korea). [Project Number 1545020852].

I am grateful to all members of Imagination Garden Inc. for their trust on my project.